\documentclass{article}
\usepackage{authblk}
\usepackage{booktabs}           
\usepackage{multirow}           
\usepackage{amsfonts}           
\usepackage{graphicx}           
\usepackage{duckuments}         
\usepackage{algorithm}
\usepackage{subcaption}
\usepackage[noend]{algpseudocode}
\usepackage{amsmath,amsthm,mathtools}
\newtheorem{definition}{Definition}

\usepackage[numbers,compress,sort]{natbib}

\author[1]{Congsong~Zhang}
\author[2]{Yong~Gao}
\author[3]{James~Nastos}
\affil[1,2]{Department of Computer Science, University of British Columbia Okanagan, Kelowna, BC, Canada}
\affil[3]{Department of Computer Science, Okanagan College, Kelowna, BC, Canada}
\affil[1]{congsong.zhang@ubc.ca}
\affil[2]{yong.gao@ubc.ca}
\affil[3]{JNastos@okanagan.bc.ca}
\title{Learning Branching Heuristics from Graph Neural Networks}
\date{}

\begin{document}

\maketitle

\begin{abstract}
Backtracking has been widely used for solving problems in artificial intelligence (AI), including constraint satisfaction problems and combinatorial optimization problems. Good branching heuristics can efficiently improve the performance of backtracking by helping prune the search space and leading the search to the most promising direction. In this paper, we first propose a new graph neural network (GNN) model designed using the probabilistic method. From the GNN model, we introduce an approach to learn a branching heuristic for combinatorial optimization problems. In particular, our GNN model learns appropriate probability distributions on vertices in given graphs from which the branching heuristic is extracted and used in a backtracking search. Our experimental results for the (minimum) dominating-clique problem show that this learned branching heuristic performs better than the minimum-remaining-values heuristic in terms of the number of branches of the whole search tree. Our approach introduces a new way of applying GNNs towards enhancing the classical backtracking algorithm used in AI.
\end{abstract}

\section{Introduction}
Using machine learning (ML), especially deep learning, to solve fundamental problems in artificial intelligence (AI) such as constraint satisfaction problems (CSPs) and combinatorial optimization problems (COPs) has attracted great attention. For example, ML techniques play an important role in SATzilla \cite{xu2008satzilla}, a portfolio-based Boolean-satisfaction-problem (SAT) solver that is able to select different algorithms depending on the characteristics of SAT instances. Convolutional neural networks have been applied to predict the behavior of SAT instances, see \cite{xu2018towards} and \cite{selsam2018learning}. Recently, graph neural networks (GNNs) have been used to learn a probability distribution to sample the solution space directly or learn greedy heuristics for approximation algorithms of COPs under the framework of reinforcement learning, see \cite{vinyals2015pointer}, \cite{li2018combinatorial}, \cite{khalil2017learning}, and \cite{kool2018attention}.

Our work differs from these approaches in that GNNs are used to learn branching heuristics to enhance exact-backtracking-search methods. While human-designed branching heuristics, such as the minimum-remaining-values (MRV) heuristic, have long been used in backtracking-based solvers, past research indicated that their effectiveness is problem-instance-dependent. Inspired by \cite{NEURIPS2020}, which (to our best knowledge) is the first work to introduce the probabilistic method into GNNs, we propose a new approach to apply the probabilistic method on GNNs to find subgraphs that satisfy certain combinatorial constraints in a given graph. The node features output by GNNs are interpreted as the probability distributions on vertices under which the information entropy is used in the branching heuristic.

In this paper, we focus on applying this GNN-based approach to solve the dominating-clique problem. We choose the dominating-clique problem as our current research interest for the following two reasons.
\begin{itemize}
    \item In \cite{culberson2005phase}, Culberson et al. study the phase transition of the dominating clique in $G(n,p)$\footnote{The $G(n,p)$ model constructs a graph with $n$ vertices by connecting each pair of vertices with probability $p$ independently.} and propose a solver for finding a dominating clique. An interesting observation in their empirical studies on the random instances at the phase transition is that the variance of the number of branches of the solvers is extremely small as compared with the average number of branches, suggesting that such hard-problems instances seem to resist commonly-used and human-designed branching heuristics;
    \item Previous approaches, such as the framework in \cite{NEURIPS2020}, attempt to focus on problems that the properties of solutions are defined locally, e.g. the maximum-clique problem, while the dominating clique problem is an interesting example of problems where a solution as a subset of vertices/variables has to satisfy some non-trivial additional global conditions defined by all the vertices in a graph.
\end{itemize}

This paper is organized as follows. The next section introduces the definition of dominating cliques and our probabilistic-method GNN. Section 3 discusses the details of our work. In Section 4, we provide and discuss experimental results. In the last section, we conclude with a brief discussion of potential future work.

\section{Dominating Clique and Probabilistic-Method GNN}

Given a graph $G = (V, E)$, a \textit{dominating clique} (DC) is a subset $S$ of $V$ such that $S$ is a clique and dominates $V \setminus S$, i.e. $\forall u\in V \setminus S$, $\exists v \in S$ such that $\{v,u\} \in E$. Checking the existence of a dominating clique in a graph is an NP-complete problem \cite{kratsch2017algorithms}. 

Our GNN model is similar to what has been proposed in \cite{NEURIPS2020}. Its architecture is standard and is trained in an unsupervised way to learn a probability distribution for each vertex. Unlike \cite{NEURIPS2020} where the distribution is directly used to sample a solution, we use the distribution to design branching heuristics. Due to the fact that a solution to the dominating-clique problem needs to satisfy both local and global properties, designing a good loss function is much more challenging than the those used in \cite{NEURIPS2020} for solutions that largely require certain local properties.    

We build a probability space by defining a Bernoulli distribution for each vertex of a given graph. By the correlation inequality in the probabilistic method, we can get an approximation of the probability measure of dominating cliques in the graph. Using the approximation as the loss function of a GNN model, we hope that the approximation grows larger as much as possible. We note that in \cite{NEURIPS2020}, the first-moment method is used in the loss function of their GNN model, which requires that the value of loss function after training is less than a constant depending on specific graphs. If the requirement is not met, the learned probability distributions do not guarantee the usefulness for the solving problem. Our loss function avoid such restrictions. In our approach, under the learned probability space, the information entropy of vertices is used in the new branching heuristic to replace the original branching heuristic of the dominating-clique solver in \cite{culberson2005phase}. Our experimental results show that the learned branching heuristic can prune more search space than the original branching heuristic.

A recent attempt towards learning branching choices in a combinatorial search was given by Li et al. (2018) \cite{li2018combinatorial} where they approached the maximum-independent-set problem by choosing a node that is expected to be in an optimal solution. This choice is made by assigning probabilities to the unselected nodes and choosing a node of high probability, and these probabilities are learned. In a sense, it performs a greedy selection of vertices to be added to a maximum independent set. This can be parallelized into a tree search by considering multiple probability maps of the vertices at each stage (according to a pre-chosen branching factor parameter), and continuing the selection process in each probability map. The largest independent set is found over all branches. Later, B{\"o}ther et al. (2022) \cite{bother2022whats} showed that the performance of such a tree search of selections can be matched by randomly generating values in place of the learned probability map for this combinatorial problem. This is comparable to a Monte Carlo tree search.
 
We contrast our work here to that of Li et al.'s tree search in that their search tree is inherently a heuristic search algorithm while ours is an exact backtracking solver. Li et al.'s search tree is parameterized with a pre-set branching factor $M$ (which they evaluate experimentally and suggest an ideal value of $M=32$), while our search tree branches on choices made in local structures. As we are interested in exact search, an important metric we use is in counting the branches expanded in our search tree, rather than counting the fraction of solved problems like in \cite{li2018combinatorial}.

 In the following, we denote by $N(v)$ the open neighborhood of a vertex $v$ (i.e. the subset of vertices that are adjacent to $v$) and $N[v] = N(v) \cup \{v\}$ the closed neighborhood. We also denote by $[n]$ the set of integers $\{1, 2, ..., n\}$.

\section{Branching Heuristic from New Probabilistic-Method GNN}

In this section, we show the restrictions in the design of the loss function of the GNN model in \cite{NEURIPS2020}. To avoid such restrictions, we propose a new approach to design loss functions using techniques and tools in the probabilistic method. We then demonstrate the utility of our approach in the maximum-clique problem. In the hope of enhancing exact algorithms for combinatorial optimization problems, we show a method to extract the branching heuristic from the learned probability space output by our GNN model. Then, we apply this method to the dominating-clique problem, and the experimental results show that the learned probability space yields to a better branching heuristic for the dominating-clique problem.

Our probabilistic-method GNN gives rise to a probability space similar to the way in \cite{NEURIPS2020}. In our probabilistic-method GNN, the node feature associated with each vertex is a 1-dimensional vector and is interpreted as the parameter of a Bernoulli distribution. Intuitively, this Bernoulli distribution characterizes the likelihood for the vertex to be in a solution. We also assume that the collection of Bernoulli distributions associated with the vertices are mutually independent. For a given graph $G(V, E)$, the collection of Bernoulli distributions give rise to a probability space $(\Omega, \mathcal{F}, \mathbb{P})$. The sample space $\Omega$ is the power set of $V$; the event space $\mathcal{F}$ is the power set of $\Omega$; for a subset $S$ of $V$, $\mathbb{P}(S) = \big(\prod_i p_i\big)\big (\prod_j (1-p_j) \big)$ with $v_i \in S$ and $v_j \notin S$ where $p_i$ and $p_j$ are the the parameters of the Bernoulli distributions associated with vertices $v_i$ and $v_j$.

\subsection{New Probabilistic-Method GNN Model}
The idea of the design of loss functions in \cite{NEURIPS2020} comes from the first-moment method in the probabilistic method
$$\mathbb{P}(X < a) > 1 - \mathbb{E}(X)/a$$
where $X$ is a random variable under a certain probability distribution and $a$ is a positive number. Applying $a = \mathbb{E}(X)/(1-t)$ with a strictly positive $t < 1$, we get
$$\mathbb{P} \big (X < \mathbb{E}(X)/(1-t) \big) > t,$$
which tells us that with a strictly positive probability $t$, $X$ is smaller than $\mathbb{E}(X)/(1-t)$. Using the first-moment method, given a combinatorial optimization problem with the quality of a solution $S$ measured by a function $f$ (such as the size of $S$), Karalias et al. in \cite{NEURIPS2020} set 
$f(S)+ 1_{S \notin \mathcal{S}} \beta $
as a random variable where S is a randomly selected subset according to the distribution from the GNN output, $\mathcal{S}$ is the set of solutions, $1_{S \notin \mathcal{S}}$ is an indicator function, and $\beta$ is a positive constant strictly greater than max$_{S \in \mathcal{S}}\big(f(S)\big)$. This random variable leads the loss function of GNN models to be
$$\mathcal{L} = \mathbb{E}\big(f(S) +  1_{S \notin \mathcal{S}} \beta\big) = \mathbb{E}\big(f(S)\big) + \mathbb{P}(S \notin \mathcal{S})\beta .$$

After training, if $\mathcal{L} < (1-t)\beta$, then with a strictly positive probability $t$, there is an $S$ with $$f(S)+ 1_{S \notin \mathcal{S}} \beta < \beta$$
which implies that $1_{S \notin \mathcal{S}}$ is false, i.e. $S \in \mathcal{S}$ and $f(S) < \beta$.

This approach is quite interesting, but it suffers from the following two drawbacks.
\begin{itemize}
    \item We want $t$ to be close to 0 so that $\mathcal{L} < (1-t)\beta$ has more likelihood after training, but it also implies the less likelihood (i.e. the probability $t$) of the event that there is an $S$ with $f(S)+ 1_{S \notin \mathcal{S}} \beta < \beta$;
    \item Choosing the value of $\beta$ needs special care. Since it is usually hard to get the closed form of $\mathbb{P}(S \notin \mathcal{S})$ for the given combinatorial optimization problem, we use an upper bound of $\mathbb{P}(S \notin \mathcal{S})$ to replace it in the loss function, which makes less likely that the loss function will converge to a small number after training if $\beta$ is too great. 
\end{itemize}

To avoid these issues, we propose a new way to design loss functions. Our idea is straightforward: minimizing $\mathbb{E}\big(f(S)\big)\mathbb{P}(S \in \mathcal{S})^{-1}$ to let the probability measure of the solutions with low values of $f$ be as great as possible. Our loss function is  
$$\ln\Big(\mathbb{E}\big(f(S)\big)\Big) - \ln\big(\mathbb{P}(S \in \mathcal{S})\big).$$
In practice, the difference between $\mathbb{E}\big(f(S)\big)$ and $\mathbb{P}(S \in \mathcal{S})$ might be large, which may cause the loss function to weigh $\mathbb{E}\big(f(S)\big)$ too strongly compared to $\mathbb{P}(S \in \mathcal{S})$. We use $\ln(\cdot)$ in hopes that the loss function would weigh the terms as proportionally as possible.

\subsection{New Loss Function for Maximum-Clique Problem}
In this section, we show an application of our GNN model for the maximum-clique problem. For the probability $\mathbb{P}(S \texttt { is a clique})$, we use the correlation inequality in the probabilistic method to get its lower bound. 

\begin{definition}\cite{Alon00theprobabilistic}
Let $M$ be a finite universal set $\{x_1,x_2,...,x_n\}$ and $R$ be a random subset of $M$ sampled by $\mathbb{P}(x_i \in R) = p_i$. Suppose these samples are mutually independent. An event $\mathcal{A}$ is a collection of subsets of $M$. $\mathcal{A}$ is an \textit{increasing event} if a set $S$ is in $\mathcal{A}$ implies that every superset of $S$ is in $\mathcal{A}$. Similarly, $\mathcal{A}$ is a \textit{decreasing event} if a set $S$ is in $\mathcal{A}$ implies that every subset of $S$ is in $\mathcal{A}$.
\end{definition}

Given two increasing events $\mathcal{A}$ and $\mathcal{B}$ and two decreasing events $\mathcal{C}$ and $\mathcal{D}$, the correlation inequality \cite{Alon00theprobabilistic} shows that 
\begin{equation}\label{correlation_inequality}
\begin{gathered}
\mathbb{P}(\mathcal{A}\cap \mathcal{B}) \geq \mathbb{P}(\mathcal{A}) \cdot \mathbb{P}(\mathcal{B}), \\
\mathbb{P}(\mathcal{C}\cap \mathcal{D}) \geq \mathbb{P}(\mathcal{C}) \cdot \mathbb{P}(\mathcal{D}), \\
\mathbb{P}(\mathcal{A}\cap \mathcal{C}) \leq \mathbb{P}(\mathcal{A}) \cdot \mathbb{P}(\mathcal{C}).
\end{gathered}
\end{equation}

By induction, if $\{\mathcal{A}_i\}_{i \in [n]}$ are all increasing or all decreasing events, 
\begin{equation}\label{correlation_inequality_multiple}
\begin{gathered}
\mathbb{P}(\bigwedge_{i \in [n]} \mathcal{A}_i) \geq \prod_{i \in [n]} \mathbb{P}(\mathcal{A}_i). 
\end{gathered}
\end{equation}

Given a graph $G(V, E)$, denote by $C_S$ the event that $S$ is a clique. For a non-adjacent pair of vertices $v_i$ and $v_j$, denote by $B_{ij}$ the event that $v_i$ and $v_j$ are both in $S$, so $\mathbb{P}(B_{ij}) = p_i p_j$. Clearly, $C_S$ is equivalent to $\bigwedge_{i,j} \overline{B_{ij}}$, and $\overline{B_{ij}}$'s are decreasing events. Thus, by (\ref{correlation_inequality_multiple}),
\begin{equation}\label{C_S}
\begin{gathered}
\mathbb{P}(C_S) = \mathbb{P}(\bigwedge_{\{v_i, v_j\} \notin E} \overline{B_{ij}}) \geq \prod_{\{v_i, v_j\} \notin E} \mathbb{P}(\overline{B_{ij}}).
\end{gathered}
\end{equation}

Clearly, maximizing $\prod_{\{v_i, v_j\} \notin E} \mathbb{P}(\overline{B_{ij}})$ helps in maximizing $\mathbb{P}(C_S)$. Hence, our loss function for the maximum-clique problem is
\begin{equation}\label{loss:max_clique}
\begin{gathered}
\mathcal{L} = -\ln \big(\mathbb{E}(|S|) \big) - \sum_{\{v_i, v_j\} \notin E} \ln \big (\mathbb{P}(\overline{B_{ij}}) \big).
\end{gathered}
\end{equation}

We apply the probability distributions output from the GNN model using the loss function (\ref{loss:max_clique}) to the greedy approximation algorithm for the maximum-clique problem in \cite{NEURIPS2020}. See Table~\ref{tab:maximum_clique} for the experimental results. 

\subsection{Loss Function for (Minimum) Dominating-Clique Problem}

The maximum-clique problem belongs to the kind of combinatorial problems that the properties of solutions are defined locally. We would like to turn our focus towards solving combinatorial problems whose solutions have both local and global conditions. Also, previous studies of applying GNNs for combinatorial optimization problems usually use the probability distributions from GNN models for greedy approximation algorithms. Thus, we also look for a method to apply the probability distributions from GNN models to exact algorithms. The dominating-clique problem is a good problem to be studied since 1) it is a combinatorial optimization problem with both local and global conditions; 2) checking the existence of a dominating clique is NP-complete; 3) there is a powerful exact solver for the dominating-clique problem in \cite{culberson2005phase} to which we can apply our improvement. 

We first design a loss function for finding a dominating clique and then extend it seamlessly for finding the minimum dominating clique. Denote by $D_S$ the event that $S$ is a dominating set, i.e. $S$ dominates $V\setminus S$. The closed form of $\mathbb{P}(D_S \cap C_S)$ is a good candidate as a loss function for finding a dominating clique, but it is harder to evaluate the closed form compared to evaluating the closed form of $\mathbb{P}(C_S)$. Thus, we alternatively analyze the upper and lower bounds of $\mathbb{P}(D_S \cap C_S)$. Our idea is to increase the upper and lower bounds of $\mathbb{P}(D_S \cap C_S)$ simultaneously to optimize its value. For a node $v_i$, denote by $A_i$ the event that the vertices in $N[v_i]$ are all not in $S$, so $\mathbb{P}(A_i) = \prod_{v_j \in N[v_i]} (1-p_j)$. It is easy to see that $D_S$ is equivalent to $\bigwedge_i \overline{A_i}$ and $\overline{A_i}$'s are increasing events. By (\ref{correlation_inequality}) and (\ref{correlation_inequality_multiple}), we get
\begin{align*}
&\mathbb{P}(D_S \cap C_S) \leq \mathbb{P}(D_S) \mathbb{P}(C_S), \\
&\mathbb{P}(D_S) = \mathbb{P}(\bigwedge_{i \in [n]} \overline{A_i}) \geq \prod_{i \in [n]} \mathbb{P}(\overline{A_i}).
\end{align*}
With (\ref{C_S}), we have 
\begin{align*}
\exp\Big(\sum_{i \in [n]} \ln \big (\mathbb{P}(\overline{A_i}) \big ) + \sum_{\{v_i, v_j\} \notin E} \ln \big (\mathbb{P}(\overline{B_{ij}}) \big) \Big) &= \prod_{i \in [n]} \mathbb{P}(\overline{A_i}) \prod_{\{v_i, v_j\} \notin E} \mathbb{P}(\overline{B_{ij}}) \\ &\leq \mathbb{P}(D_S) \mathbb{P}(C_S).
\end{align*}

This inequality tells us that increasing 
$\sum_{i \in [n]} \ln \big (\mathbb{P}(\overline{A_i}) \big ) + \sum_{\{v_i, v_j\} \notin E} \ln \big (\mathbb{P}(\overline{B_{ij}}) \big)$
is meanwhile increasing the upper bound of $\mathbb{P}(D_S \cap C_S)$. 

For the lower bounds of $\mathbb{P}(D_S \cap C_S)$, we have
\begin{align*}
\mathbb{P}(D_S \cap C_S) &\geq \mathbb{P}(D_S) + \mathbb{P}(C_S) - 1 \\
&\geq 2\exp\Big(\frac{1}{2} \big(\ln{\mathbb{P}(D_S)} + \ln{\mathbb{P}(C_S)}\big)\Big) - 1 &&\text{(as $e^x$ is a convex function)} \\
&\geq 2\exp \Big ( \frac{1}{2} \Big ( \sum_{i \in [n]} \ln \big (\mathbb{P}(\overline{A_i}) \big ) + \sum_{\{v_i, v_j\} \notin E} \ln \big (\mathbb{P}(\overline{B_{ij}}) \big) \Big) \Big) - 1.
\end{align*}

Therefore, we set the loss function for the dominating-clique problem as 
\begin{equation}\label{loss_for_dcs}
\mathcal{L} = - \Big (\sum_{i \in [n]} \ln \big (\mathbb{P}(\overline{A_i}) \big ) + \sum_{\{v_i, v_j\} \notin E} \ln \big (\mathbb{P}(\overline{B_{ij}}) \big) \Big)
\end{equation}
trying to optimize the upper and lower bounds of $\mathbb{P}(D_S \cap C_S)$ simultaneously.

To find the minimum dominating clique, we can simply add $\ln\Big(\mathbb{E}(|S|)\Big)$ into the loss function as
\begin{equation}\label{loss_for_min_dcs}
\mathcal{L} = - \Big ( \sum_{i \in [n]} \ln \big (\mathbb{P}(\overline{A_i}) \big ) + \sum_{\{v_i, v_j\} \notin E} \ln \big (\mathbb{P}(\overline{B_{ij}}) \big) \Big) + \ln \Big (\mathbb{E}(|S|) \Big).
\end{equation}

By Equation (\ref{loss_for_min_dcs}), we make an effort to increase the probability measure of small-sized dominating cliques so that it is easier to identify them. However, calculating $\mathbb{E}(|S|)$ on $\Omega$ (i.e. $\mathbb{E}(|S|) = \sum_{i=1}^{n} p_i$) indeed affects all small-sized subsets of $V$ rather than small-sized dominating cliques. An improvement is to find an event in the event space $\mathcal{F}$ which is close to the exact event consisting of dominating cliques only. One way to accomplish this is as follows. 

In each iteration during the GNN training, we generate a random permutation $\{v_1,v_2,...,v_n\}$ of $V$ and generate an event (i.e. a set of subset of $V$) iteratively. We initialize the event as an empty set. For $i = 1$ to $n$, we first exclude the subsets of $V$ that contain any vertex of $v_1,...,v_{i-1}$. Then, we continue to exclude the subsets of $V$ that do not contain $v_i$. After that, we exclude the subsets of $V$ that contain any non-adjacent vertex to $v_i$. At the end of the current iteration, we add the remaining subsets of $V$ into the event. 

This event is much more closer to the event of dominating cliques only than the event $\Omega$. Thus, instead of minimizing $\mathbb{E}(|S|)$ on $\Omega$, we try to minimize $\mathbb{E}(|S|)$ on this event as
\begin{equation}\label{subspace_expected}
\sum_{i=1}^{n} \Big (p_i \prod_{j=1}^{i-1} (1-p_j) \prod_{v_k \in \{v_{i+1},..,v_{n}\} \setminus N(v_i)} (1-p_k) \big(1 + \sum_{v_r \in N(v_i) \setminus \{v_1,..,v_{i-1}\}} p_r \big) \Big)
\end{equation}
in hopes that it gives the loss function a more accurate expected size of dominating cliques. The term $p_i$ is the probability of the event that $v_i$ is in $S$. The term $\prod_{j=1}^{i-1} (1-p_j)$ is the probability of the event that $v_1, ..., v_{i-1}$ are not in $S$. The term $\prod_{v_k \in \{v_{i+1},..,v_{n}\} \setminus N(v_i)} (1-p_k)$ is the probability of the event that $S$ do not contain any vertex that is behind $v_i$ in the permutation and non-adjacent to $v_i$. The term $\big(1 + \sum_{v_r \in N(v_i) \setminus \{v_1,..,v_{i-1}\}} p_r\big)$ is the conditional expectation of $|S|$ given the above three events happen. 

\subsection{New Branching Heuristic for Dominating-Clique Solvers}
Exact solvers for combinatorial optimization problems are generally based on backtracking search. To improve this framework, branching heuristics are applied to give the most promising direction during the search, and branching heuristics are designed by the properties of specific problems. To improve branching heuristics by our GNN model, our idea is to define a function $f(\text{Var$_b$}, \Theta)$ to measure the quality of branches in the search tree where Var$_b$ is the set of unsigned variables of a branch $b$ and $\Theta$ is the probability space from the output of our GNN model. We next utilize this idea for the dominating-clique problem. 

Culberson et al. \cite{culberson2005phase} propose an efficient solver for the dominating-clique problem. From their experiments, the solver performs better than other general SAT solvers, including BerkMin, MarchEq, and SATzilla. The solver is a backtracking-based algorithm. They encode a given graph $G=(V,E)$ with $V=\{v_1,v_2,...,v_n\}$ to a CNF formula as follows.

\begin{itemize}
    \item There are $n$ variables $\{X_i\}_{i \in [n]}$ where $X_i=1$ means that the corresponding $v_i$ is in the solution;
    \item Define a clause $C_i = \{X_j\}_{v_j \in N[v_i]}$ for each vertex $v_i$ to indicate that the vertex is in the solution or at least one of its neighbors is in the solution;
\end{itemize}

With the encoded CNF formula, the solver works as Algorithm~\ref{alg:dcs_yong}.

\begin{algorithm}[H]
\caption{\cite{culberson2005phase} An algorithm for the dominating-clique problem}
\label{alg:dcs_yong}
\begin{algorithmic}[1]
\Procedure{DomClq}{$D, S, U, G(V,E)$}
\If{$U$ is $\emptyset$}
    \State $DOMCLQ \gets D$
    \State \Return{} TRUE
\Else
    \State Find $C \in U$ such that $|C \cap S| = \min_{C' \in U}|C' \cap S|$
    \If{$C \cap S$ is not $\emptyset$}
        \State $S'' \gets S$
        \For{$X_i \in C \cap S$}
            \State $X_i \gets 1$; $D \gets D \cup \{v_i\}$ 
            \State $S' \gets \{X_j \text{ } | \text{ } X_j \in S'', \{v_i, v_j\} \in E\}$; $U' \gets U \setminus \{C' \text{ } | \text{ } C' \in U, X_i \in C'\}$
            \If{DomClq($D, S', U', G(V,E)$)}
                \State \Return{} TRUE
            \Else
                \State $X_i \gets 0$; $D \gets D \setminus \{v_i\}$; $S'' \gets S'' \setminus \{X_i\}$
            \EndIf
        \EndFor
    \EndIf
    \State \Return{} FALSE
\EndIf
\EndProcedure
\end{algorithmic}
\end{algorithm}

The solver has three parameters: a potential dominating clique $D$, a set $S$ of the unassigned variables such that $S = \{X_u \text{ }| \text{ } \forall v \in D, \{u,v\} \in E \}$, and a set $U$ of unsatisfied clauses. The three parameters are initialized as $\emptyset$, $\{X_i\}_{i \in [n]}$, and $\{C_i\}_{i \in [n]}$ respectively. 

From the perspective of CSP solvers, this algorithm uses the MRV heuristic. This heuristic is optimal if every sub-problem by adding one vertex into $D$ has the same amount of search space. However, it is false for most cases. To improve it, we can see $\{X_i\}_{i \in [n]}$ as random variables under the probability distributions output from our probabilistic-method GNN. We use the information entropy of these random variables to predict the amount of the search space of unsatisfied clauses. In particular, instead of the cardinality of $C' \cap S$, we measure the joint information entropy of the random variables in $C' \cap S$. We therefore replace Line 6 in Algorithm~\ref{alg:dcs_yong} by finding $C \in U$ such that $(C \cap S)$ has the minimum joint information entropy. In other words, we define the function $f$ as the joint information entropy of unassigned variables of a branch to measure the quality of branches. Note that the probability distributions of unassigned variables should not be fixed during the backtracking search. The reason is that $S$, $D$, and $U$ are changing during the search, so the correspondingly unexplored subgraph which consists of $S$ and $U$ is also changing. We apply the softmax function on the distributions of the unassigned variables to re-weigh them during the backtracking search. We also try the $Z$-score normalization, but its performance is worse than the softmax function. 

To calculate the joint information entropy of random variables in $C' \cap S = \{X_1, ..., X_m\}$, we try two ways. The first one, called the \textbf{fast} version because the calculation of this way is fast but not accurate, is $\sum_{i=1}^{m} \mathbb{H}(X_i) - \Big ( \prod_{i=1}^{m}(1-p_i) \Big ) \log_2 \Big (\prod_{i=1}^{m}(1-p_i) \Big )$. $X_i$'s are mutually independent, so their joint information entropy is the sum of each one's information entropy. In addition, if there is a dominating clique, at least one variable in $C' \cap S$ is 1; we thus minus the information entropy of the case that all variables in $C' \cap S$ are 0. 

The second way, called the \textbf{accurate} version but with a slow calculation, is similar to the improvement on Equation (\ref{loss_for_min_dcs}). Instead of calculating the joint information entropy of random variables in $(C' \cap S)$, we calculate the joint information entropy under a more precise case. Given a clause $C' \cap S = \{X_1,..,X_m\}$, for $i = 1$ to $m$, we iteratively set $X_1, ..., X_{i-1}$ as 0 and $X_i$ as 1. We know that $\{X_{i+1}, ..., X_m\} \setminus \{X_j\}_{v_j \in N(v_i)}$ must be 0 from the properties of dominating cliques. Thus, we only need to consider the joint information entropy of the random variables in $N[v_i] \setminus \{v_1, ..., v_{i-1}\}$. Similar to Equation (\ref{subspace_expected}), we calculate the joint information entropy as 
\begin{equation}\label{subspace_entropy}
\sum_{i=1}^{m} \Big (p \big(-\log_2(p) + \sum_{v_r \in N(v_i) \setminus \{v_1,..,v_{i-1}\}} \mathbb{H}(X_r) \big) \Big)
\end{equation}
where $p = p_i \prod_{j=1}^{i-1} (1-p_j) \prod_{v_k \in \{v_{i+1},..,v_{m}\} \setminus N(v_i)} (1-p_k)$.

\section{Experimental Results}

\subsection{Finding Maximum Cliques}
Table~\ref{tab:maximum_clique} shows the experimental results of approximation ratio (i.e. the ratio of the size of the clique we find versus the size of the maximum clique). We use the greedy approximation algorithm in \cite{NEURIPS2020} to find large cliques, and the maximum clique is from Gurobi \cite{gurobi}. The greedy approximation algorithm has two ways, fast and slow, to decode cliques using the learned probability distributions. The fast way returns quickly but with a less use of the input probability distributions compared to the slow way. We run experiments for two GNN models. These two models have the same neural network architecture as the model in \cite{NEURIPS2020}. The only difference is that one uses our loss function (\ref{loss:max_clique}) and another uses the original loss function in \cite{NEURIPS2020}. The datasets Twitter, COLLAB, and IMDB used in the experiments are from \cite{NEURIPS2020} as well.

\begin{table}
    \center
    \caption{Results on test dataset: approximation ratios on real-world datasets and $G(n,p)$ instances; the average number of nodes of $G(n,p)$ instances is similar to the average number of nodes of graphs in Twitter (about 120 nodes per graph); Sparse means that the instances are generated with $0.2 \leq p \leq 0.4$; Dense means that the instances are generated with $0.6 \leq p \leq 0.8$; italics indicates the standard deviation.}
    \label{tab:maximum_clique}
    \resizebox{1\textwidth}{!}{
    \begin{tabular}{ccccc}
        \toprule
        \multicolumn{1}{c}{} & \multicolumn{2}{c}{Loss function as (\ref{loss:max_clique})} & \multicolumn{2}{c}{Loss function in \cite{NEURIPS2020}} \\
        \cmidrule(lr){1-1} \cmidrule(lr){2-3} \cmidrule(lr){4-5}
        \(\text{Dataset}\) & \(\text{Slow}\) & \(\text{Fast}\) & \(\text{Slow}\) & \(\text{Fast}\) \\
        \midrule
        Twitter & 0.931 $\pm$ \emph{0.099} & 0.927 $\pm$ \emph{0.112} & 0.956 $\pm$ \emph{0.077} & 0.914 $\pm$ \emph{0.141} \\
        COLLAB & 1.000 $\pm$ \emph{0.000} & 1.000 $\pm$ \emph{0.000} & 1.000 $\pm$ \emph{0.000} & 1.000 $\pm$ \emph{0.000} \\
        IMDB & 1.000 $\pm$ \emph{0.000} & 1.000 $\pm$ \emph{0.000} & 1.000 $\pm$ \emph{0.000} & 1.000 $\pm$ \emph{0.000} \\
        $G(n,p)$(Sparse) & 0.910 $\pm$ \emph{0.100} & 0.908 $\pm$ \emph{0.102} & 0.922 $\pm$ \emph{0.099} & 0.901 $\pm$ \emph{0.104} \\
        $G(n,p)$(Dense) & 0.901 $\pm$ \emph{0.050} & 0.900 $\pm$ \emph{0.047} & 0.886 $\pm$ \emph{0.062} & 0.893 $\pm$ \emph{0.053} \\
        \bottomrule
    \end{tabular}}
\end{table}

\subsection{Finding Dominating Cliques}
We run experiments on three dominating-clique solvers: the original solver in \cite{culberson2005phase} and the GNN-enhanced solvers (the output distributions from the GNN model with the loss function (\ref{loss_for_dcs}) as input) using the fast and accurate calculations of joint information entropy respectively.

\begin{figure}[H]
    \begin{subfigure}{.5\textwidth}
        \centering
        \includegraphics[width=.8\linewidth]{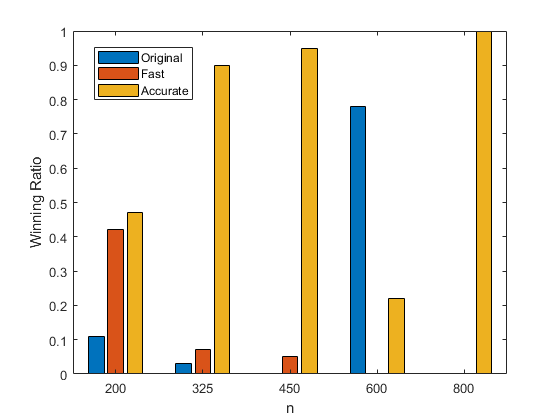}
        \caption{}
        \label{fig:dc_unsat_winning_ratio}
    \end{subfigure}%
    \begin{subfigure}{.5\textwidth}
        \centering
        \includegraphics[width=.8\linewidth]{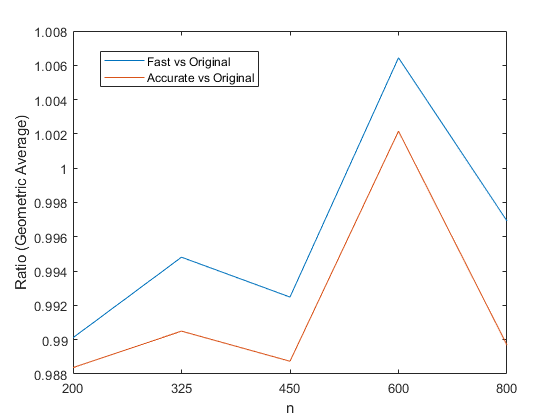}
        \caption{}
        \label{fig:dc_unsat_avg_ratio}
    \end{subfigure} \\
    \begin{subfigure}{.5\textwidth}
        \centering
        \includegraphics[width=.8\linewidth]{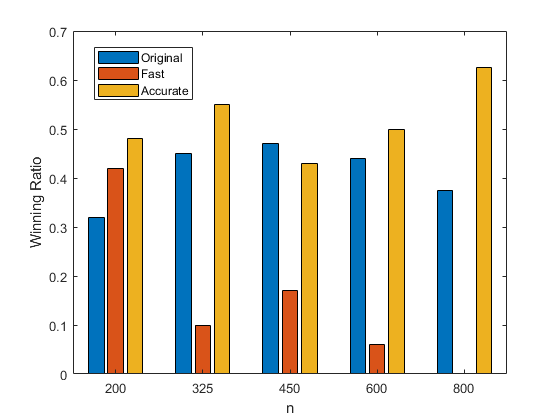}
        \caption{}
        \label{fig:dc_sat_winning_ratio}
    \end{subfigure}%
    \begin{subfigure}{.5\textwidth}
        \centering
        \includegraphics[width=.8\linewidth]{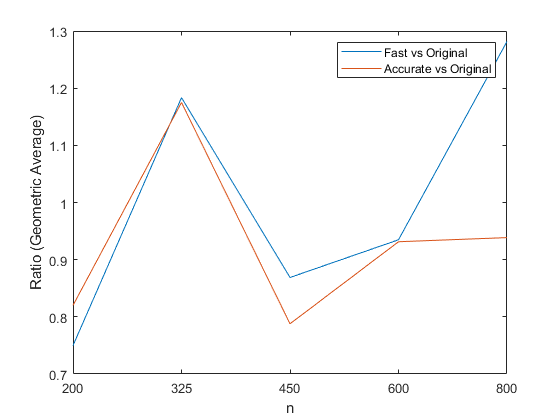}
        \caption{}
        \label{fig:dc_sat_avg_ratio}
    \end{subfigure}
\caption{Experimental results on test data for finding dominating cliques; X axis for $n$}
\label{dc}
\end{figure}

Figure~\ref{dc} shows the experimental results on test data. To be specific, Figure~\ref{fig:dc_unsat_winning_ratio} shows the winning ratios of the three solvers on the instances that dominating cliques do not exist, and Figure~\ref{fig:dc_sat_winning_ratio} shows the winning ratios of the three solvers on the instances that dominating cliques exist. Figure~\ref{fig:dc_unsat_avg_ratio} shows the ratios of the GNN-enhanced solvers to the original algorithm in terms of geometric average number\footnote{The reason of using the geometric average is that the range of the number of branches from the experimental results is big, especially for the instances that dominating cliques exist. Under this case, the arithmetic average may not reveal the performance of solvers correctly.} of their branches on the instances that dominating cliques do not exist. Figure~\ref{fig:dc_sat_avg_ratio} shows such ratios on the instances that dominating cliques exist. It is reasonable that the GNN-enhanced solvers perform better on the instances that dominating cliques do not exist. For such case, solvers go through the whole search tree, so the learned branching heuristic can prune more search space. Instead, for the case of solving instances that dominating cliques exist, the original heuristic might be equally effective since solvers return once a dominating clique is found. 

\subsection{Finding Minimum Dominating Cliques}
To find the minimum dominating clique, we add the following features to our dominating-clique solvers. We record the currently minimum dominating clique and update it once we find a dominating clique with lower cardinality, and so on until completing the whole backtracking search. In addition, we apply backjumping to prune the search space. Our evaluation order of the random variables in $C \cap S$ is the decreasing order of the numbers of the unsatisfied clauses that these random variables involve. Thus, we can backjump three levels whenever we get the first dominating clique or find a dominating clique with lower cardinality than the currently minimum solution. Also, we can backjump two levels whenever the current recursive depth is equal to the cardinality of the currently minimum dominating clique. 

\begin{figure}[H]
    \begin{subfigure}{.5\textwidth}
        \centering
        \includegraphics[width=.8\linewidth]{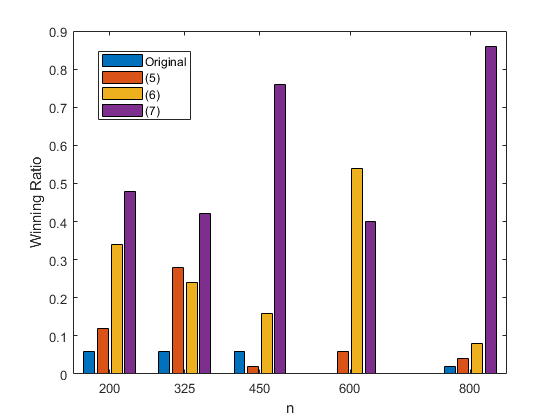}
        \caption{}
        \label{fig:mdc_fast}
    \end{subfigure}%
    \begin{subfigure}{.5\textwidth}
        \centering
        \includegraphics[width=.8\linewidth]{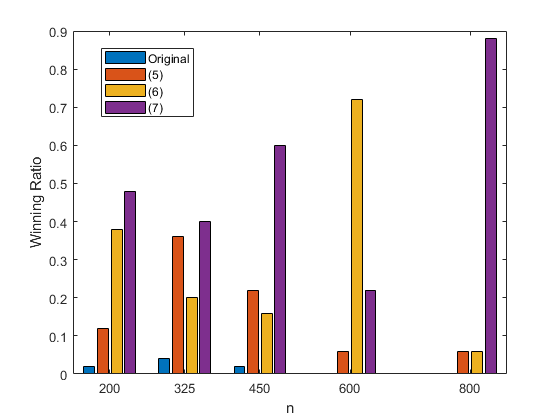}
        \caption{}
        \label{fig:mdc_accurate}
    \end{subfigure} \\
    \begin{subfigure}{.8\textwidth}
        \centering
        \includegraphics[width=1.\linewidth]{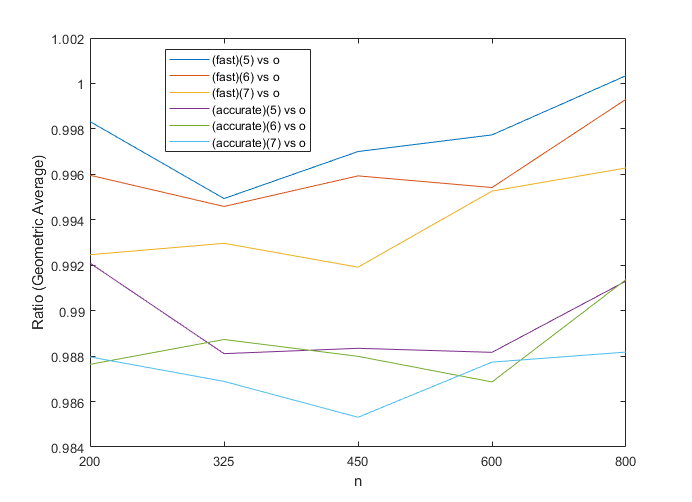}
        \caption{}
        \label{fig:mdc_ratio}
    \end{subfigure}%
    
\caption{Experimental results on test data for finding minimum dominating cliques; X axis for $n$}
\label{mdc}
\end{figure}

We prepare three GNN models for the experiments. Their loss functions are (\ref{loss_for_dcs}), (\ref{loss_for_min_dcs}) with $\mathbb{E}(|S|) = \sum_{i=1}^{n} p_i$, and (\ref{loss_for_min_dcs}) with $\mathbb{E}(|S|)$ as (\ref{subspace_expected}) respectively. Combined with the two calculations of joint information entropy, we have six solvers. Figure~\ref{mdc} shows the experimental results on test data for finding the minimum dominating clique. In particular, Figure~\ref{fig:mdc_fast} shows the winning ratios of the four solvers: the original solver and the other three GNN-enhanced solvers using the fast calculation of joint information entropy.
We use (\ref{loss_for_dcs}) in the figure to represent the GNN-enhanced solver using the output probability distributions from the GNN model with the loss function (\ref{loss_for_dcs}). The (\ref{loss_for_min_dcs}) in the figure represents the GNN-enhanced solver using the output probability distributions from the GNN model with the loss function (\ref{loss_for_min_dcs}) where $\mathbb{E}(|S|) = \sum_{i=1}^{n} p_i$. The (\ref{subspace_expected}) in the figure represents the GNN-enhanced solver using the output probability distributions from the GNN model with the loss function (\ref{loss_for_min_dcs}) where $\mathbb{E}(|S|)$ as (\ref{subspace_expected}). Figure~\ref{fig:mdc_accurate} shows the winning ratios of these four solvers using the accurate calculation of joint information entropy. Figure~\ref{fig:mdc_ratio} shows the ratios of these six GNN-enhanced solvers to the original solver in terms of the geometric average of branches. From Figure~\ref{fig:mdc_fast} and Figure~\ref{fig:mdc_accurate}, we can see that the loss functions that have the term $E(|S|)$ performs better than the loss function (\ref{loss_for_dcs}). Another observation from Figure~\ref{fig:mdc_ratio} is that either the accurate calculation of joint information entropy or the accurate calculation of $\mathbb{E}(|S|)$ improves the solver's performance.

\section{Concluding Remarks}
In this work, we have developed an approach that uses the probabilistic method in the design of GNN models to learn branching heuristics to enhance exact solvers for hard combinatorial problems. This approach can be applied to other COPs of finding an optimal constrained vertex subset of a given graph, such as the maximum/minimum dominating-independent-set problem or the maximum-clique problem. In the future, we would like to enhance this approach to solve the problems that ask for an optimal collection of vertex subsets where a vertex is allowed to be in multiple subsets, e.g. the tree-decomposition problem and the edge-clique-cover problem. We hope that our approach can be a starting effort of applying GNNs techniques to improve classical AI algorithms.

\section{Acknowledgement}
This research was supported in part through computational resources and services provided by UBC Advanced Research Computing, ``UBC ARC Sockeye." UBC Advanced Research Computing, 2022, doi: 10.14288/SOCKEYE. 

\bibliographystyle{unsrtnat}
\bibliography{reference}

\appendix

\section{Neural Network Architecture of GNN Model for Dominating Cliques}
We implement our GNN model by PyTorch 1.9.1 and PyTorch Geometric 2.0.1. We set the node features of an input graph as 1-dimensional vectors. As shown above, we interpret these features as the Bernoulli distributions associated with vertices. Our GNN model consists of 8 layers: 6 graph-isomorphism-network (GIN) layers and 2 linear layers followed. The reason of using GINs is that the properties of sub-structures of the graphs containing dominating cliques are similar. For example, the diameters of the graphs containing dominating cliques are at most 3, and they all have a clique as a kernel. GIN is a good option here because it has the most powerful ability in GNN layers to detect isomorphic subgraphs \cite{xu2018how}. We apply the batch normalization on the output of each layer as it is a standard operation in GNNs, see \cite{pmlr-v37-ioffe15}. In addition, we set the batch size as 32 because this number has a good practice in real applications \cite{bengio2012practical}. Furthermore, our dominating-clique solvers are implemented by the programming language C and run on Ubuntu 20.04.4 LTS with Intel CPU i7-11700F and 48GB RAM. 

\section{Preliminary Test for Dominating Cliques}
\begin{figure}[H]
\centerline{\includegraphics[width=1.\textwidth]{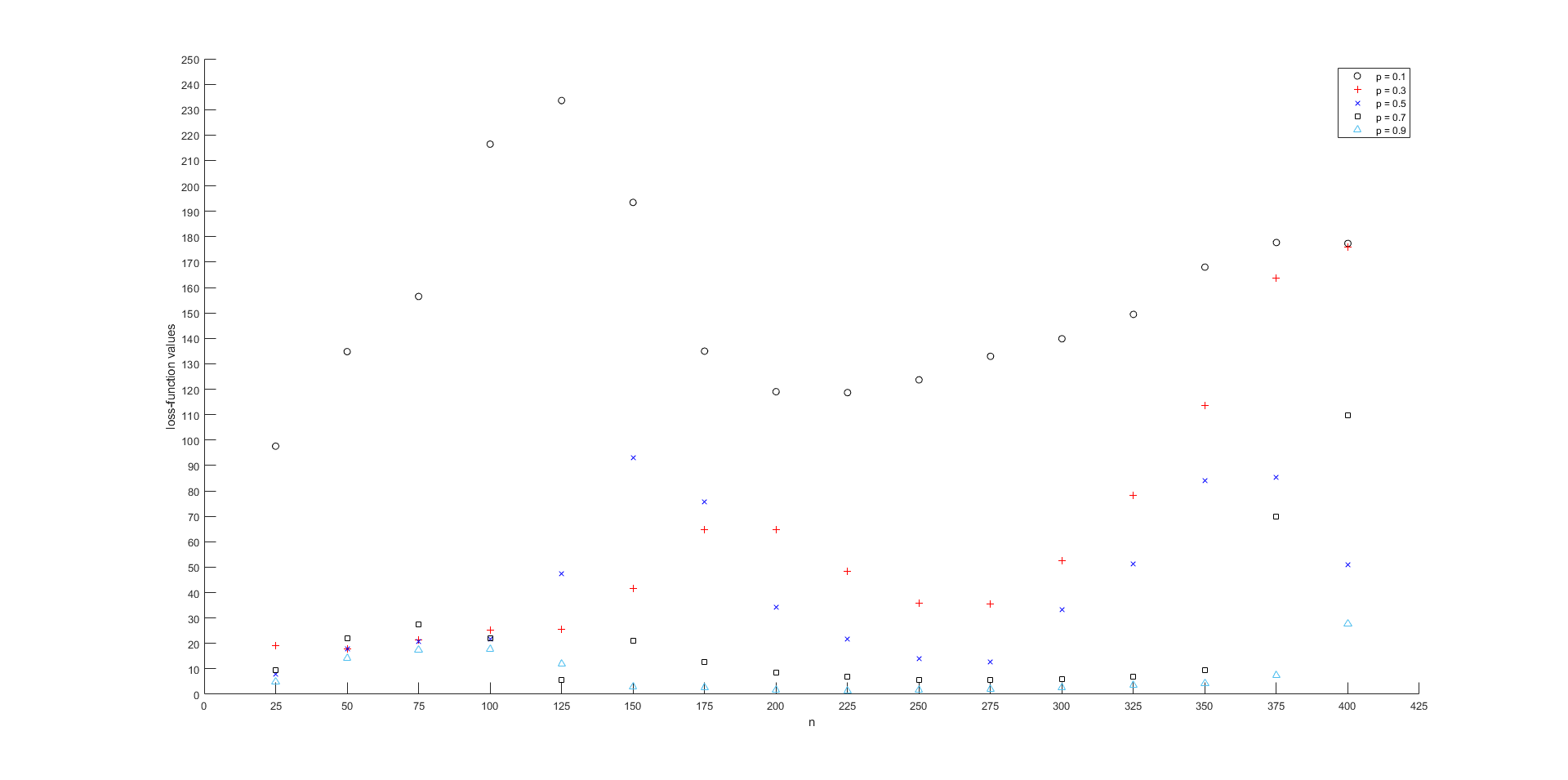}}
\caption{Averages of the values of loss function of 32 instances for each pair of $n$ and $p$; the x-axis is the size of graphs; the y-axis is the average of the values of loss function.}
\label{fig:Verification}
\end{figure}

To check whether Equation (\ref{loss_for_dcs}) works effectively as the loss function of the GNN model for finding dominating cliques, we create other $G(n,p)$ instances with different $n$ and $p$. In particular, $n$ ranges from 25 to 400 in increments of 25, and $p$ ranges from 0.1 to 0.9 in increments of 0.2, so we have 80 pairs of $n$ and $p$. Then we generate 32 instances from $G(n,p)$ for each pair of $n$ and $p$ and check the values of loss function of these instances output from the well-trained probabilistic-method GNN model. From the above phase transition, the instances generated from $G(n,p)$ where $p = 0.1$ have low likelihood of containing a dominating clique; on the contrary, the instances generated from $G(n,p)$ where $p = 0.9$ are more likely to have dominating cliques. Figure~\ref{fig:Verification} shows the average of the values of loss function of the 32 instances for each pair of $n$ and $p$. From Figure~\ref{fig:Verification}, we can see that the average of the values of loss function of the instances generated from $G(n,p)$ where $p = 0.1$ are much greater than the corresponding average of the instances generated from $G(n,p)$ where $p = 0.9$, which is consistent with the known phase transition because the greater loss-function values imply the lower probabilities of existing a dominating clique.

\section{Training Data for (Minimum) Dominating Cliques}
Our data is generated from $G(n,p)$. The work in \cite{culberson2005phase} shows that the phase transition of dominating cliques happens at the exact threshold $p = \frac{3-\sqrt{5}}{2} \approx 0.381$. It is also mentioned in \cite{culberson2005phase} that $p$ around $0.371$ is a good empirical probability to create the instances that it is hard to predict the existence of dominating cliques. We call such instances as \textbf{hard instances}. 

For learning a branching heuristic to find a dominating clique, our training data contains 2650 instances generated from $G(n,p)$ with the range of $n$ being from 75 to 800. Among these instances, 1250 instances are generated from $G(n,p)$ with $p$ being around 0.371 as hard instances. The remaining 1400 instances are generated from $G(n,p)$ with $p \in (0, 0.35) \cup (0.4, 1)$. we call them \textbf{easy instances} as they are generated outside the phase transition and relatively easier to be solved. We use 750 hard instances and all easy instances as our training data. The remaining 500 hard instances are used as validation data and test data. 

For learning a branching heuristic to find the \textbf{minimum} dominating clique, we create another 1250 instances (750 instances are used in training data, and the remaining 500 instances are used as validation data and test data) from $G(n,p)$ to replace the hard instances. The range of $n$ of these new instances is the same as the hard instances above, but the range of $p$ is between $0.4$ and $0.41$. The reason is that if a graph is dense, there are many dominating cliques with minimum size. If so, commonly used branching heuristics are powerful enough because the first-found solution is usually the minimum solution. Therefore, we choose this range of $p$ to generate the instances such that they have some dominating cliques but only few of these dominating cliques with minimum size. Additionally, we keep the same easy instances above in the training data for finding the minimum dominating clique.

\section{Raw Experimental Results for Finding Dominating Cliques}

We run three solvers, the original solver in \cite{culberson2005phase} (abbreviated as O) and the other two GNN-enhanced solvers with the fast and accurate calculations for joint information entropy (abbreviated as F and A) respectively on the training, validation, and test data where $p$ is around 0.371. We compare the number of branches of their search trees. Table~\ref{tab:cmp_training_dataset_for_exist} shows the results on the training data. Tables~\ref{tab:cmp_validation_dataset_for_exist} and~\ref{tab:cmp_test_dataset_for_exist} show the results on the validation and test data respectively. The columns O vs F, O vs A, and A vs F record the number of times that the corresponding solver wins. The Avg columns are the arithmetic average of the number of branches. The numbers in brackets are the results of the instances that dominating cliques exist.

\begin{table}[H]
    \scriptsize
    \center
    \caption{Results on Training Data (DC): 50 instances for each $n$.}
    \label{tab:cmp_training_dataset_for_exist}
    \resizebox{1\textwidth}{!}{
    \begin{tabular}{cccccccclll}
        \toprule
        \multicolumn{2}{c}{$G(n,p)$} & \multicolumn{2}{c}{O vs F} & \multicolumn{2}{c}{O vs A} & \multicolumn{2}{c}{F vs A} & \multicolumn{3}{c}{Avg} \\
        \cmidrule(lr){1-2} \cmidrule(lr){3-4} \cmidrule(lr){5-6} \cmidrule(lr){7-8} \cmidrule(lr){9-11}
        \(\text{$n$}\) & \(\text{$p$}\) & \(\mathbf{O}\) & \(\mathbf{F}\) & \(\mathbf{O}\) & \(\mathbf{A}\) & \(\mathbf{F}\) & \(\mathbf{A}\) & \(\mathbf{O}\) & \(\mathbf{F}\) & \(\mathbf{A}\) \\
        \midrule
        75 & 0.3698  & 27 (13) & 23 (10) & 26 (13) & 24 (10) & 12 (3) & 38 (20) & 97.82 (35.7) & 101.74 (44.17) & 102.6 (46.04) \\
        150 & 0.3663  & 18 (10) & 32 (16) & 17 (10) & 33 (16) & 16 (9) & 34 (17) & 774.86 (307.15) & 755.1 (279.65) & 785.42 (343.81) \\
        225 & 0.368  & 25 (16) & 25 (11) & 18 (16) & 32 (11) & 12 (10) & 38 (17) & 3320.44 (1151.48) & 3333.22 (1190.11) & 3335.72 (1224.78) \\
        250 & 0.3669  & 16 (10) & 34 (15) & 17 (11) & 33 (14) & 13 (9) & 37 (16) & 5917 (3062.56) & 5762.3 (2815.04) & 5876.2 (3080.96) \\
        300 & 0.3674  & 15 (9) & 35 (8) & 13 (8) & 37 (9) & 10 (5) & 40 (12) & 14592.46 (5661.06) & 14584.48 (5857.94) & 14366.88 (5342.59) \\
        350 & 0.3671  & 13 (10) & 37 (12) & 11 (10) & 39 (12) & 11 (8) & 39 (14) & 23925.5 (7330.55) & 24269.86 (8431.23) & 24070.1 (8105.41) \\
        375 & 0.3725  & 16 (15) & 34 (21) & 15 (14) & 35 (22) & 13 (12) & 37 (24) & 24416.82 (10556.17) & 23439.86 (9307.83) & 23376.08 (9291.94) \\
        425 & 0.3685  & 11 (9) & 39 (15) & 11 (9) & 39 (15) & 8 (8) & 42 (16) & 53869.16 (13055.96) & 53672.14 (13066.25) & 53347.44 (12800.75) \\
        475 & 0.3663  & 26 (8) & 24 (10) & 19 (9) & 31 (9) & 4 (3) & 46 (15) & 99739.56 (24710.89) & 99646.66 (24120.44) & 100467.5 (27396) \\
        525 & 0.366  & 24 (9) & 26 (10) & 21 (10) & 29 (9) & 4 (4) & 46 (15) & 149029.76 (36525.53) & 149308.26 (36148.21) & 148783.1 (36028.47) \\
        575 & 0.368  & 35 (10) & 15 (13) & 15 (6) & 35 (17) & 1 (1) & 49 (22) & 239032.36 (91339.04) & 246610.04 (104490.87) & 236344.26 (85521.7) \\
        625 & 0.3669  & 10 (6) & 40 (8) & 6 (6) & 44 (8) & 2 (2) & 48 (12) & 398240.3 (105061.86) & 398714.52 (115036.64) & 396739.72 (114490.5) \\
        675 & 0.3685  & 10 (10) & 40 (10) & 9 (9) & 41 (11) & 0 (0) & 50 (20) & 551684.76 (172927.9) & 549085.24 (171353.65) & 545489.22 (169937.75) \\
        700 & 0.3671  & 6 (6) & 44 (11) & 5 (5) & 45 (12) & 1 (1) & 49 (16) & 657604.96 (155241.12) & 654205.9 (152510.88) & 640940.52 (124779.12) \\
        750 & 0.368  & 9 (8) & 41 (5) & 8 (7) & 42 (6) & 0 (0) & 50 (13) & 1129832.74 (507520.31) & 1127597.54 (508628.38) & 1120142.64 (505230.54) \\
        \bottomrule
    \end{tabular}}
\end{table}

\begin{table}[H]
    \scriptsize
    \center
    \caption{Results on Validation Data (DC): 50 instances for each $n$.}
    \label{tab:cmp_validation_dataset_for_exist}
    \resizebox{1\textwidth}{!}{
    \begin{tabular}{cccccccclll}
        \toprule
        \multicolumn{2}{c}{$G(n,p)$} & \multicolumn{2}{c}{O vs F} & \multicolumn{2}{c}{O vs A} & \multicolumn{2}{c}{F vs A} & \multicolumn{3}{c}{Avg} \\
        \cmidrule(lr){1-2} \cmidrule(lr){3-4} \cmidrule(lr){5-6} \cmidrule(lr){7-8} \cmidrule(lr){9-11}
        \(\text{$n$}\) & \(\text{$p$}\) & \(\mathbf{O}\) & \(\mathbf{F}\) & \(\mathbf{O}\) & \(\mathbf{A}\) & \(\mathbf{F}\) & \(\mathbf{A}\) & \(\mathbf{O}\) & \(\mathbf{F}\) & \(\mathbf{A}\) \\
        \midrule
        100 & 0.3685  & 18 (11) & 32 (20) & 22 (14) & 28 (17) & 19 (10) & 31 (21) & 187.3 (74.61) & 185.14 (74.03) & 186.08 (75.19) \\
        275 & 0.3669  & 12 (6) & 38 (13) & 11 (6) & 39 (13) & 16 (9) & 34 (17) & 774.86 (307.15) & 755.1 (279.65) & 785.42 (343.81) \\
        400 & 0.3698  & 16 (14) & 34 (16) & 15 (13) & 35 (17) & 11 (10) & 39 (20) & 36675.62 (13529.03) & 36165.5 (12864.77) & 36733.5 (13968.6) \\
        550 & 0.3689  & 29 (17) & 21 (11) & 19 (17) & 31 (11) & 4 (4) & 46 (24) & 162022.64 (56828.25) & 165254.38 (62492.5) & 164349.24 (62158.89) \\
        725 & 0.3675  & 7 (7) & 43 (9) & 7 (7) & 43 (9) & 1 (1) & 49 (15) & 819762.54 (213827.06) & 804655.12 (173782.5) & 799592.72 (172590.38) \\
        \bottomrule
    \end{tabular}}
\end{table}

\begin{table}[H]
    \scriptsize
    \center
    \caption{Results on Test Data (DC): 50 instances for each $n$.}
    \label{tab:cmp_test_dataset_for_exist}
    \resizebox{1\textwidth}{!}{
    \begin{tabular}{cccccccclll}
        \toprule
        \multicolumn{2}{c}{$G(n,p)$} & \multicolumn{2}{c}{O vs F} & \multicolumn{2}{c}{O vs A} & \multicolumn{2}{c}{F vs A} & \multicolumn{3}{c}{Avg} \\
        \cmidrule(lr){1-2} \cmidrule(lr){3-4} \cmidrule(lr){5-6} \cmidrule(lr){7-8} \cmidrule(lr){9-11}
        \(\text{$n$}\) & \(\text{$p$}\) & \(\mathbf{O}\) & \(\mathbf{F}\) & \(\mathbf{O}\) & \(\mathbf{A}\) & \(\mathbf{F}\) & \(\mathbf{A}\) & \(\mathbf{O}\) & \(\mathbf{F}\) & \(\mathbf{A}\) \\
        \midrule
        200 & 0.3689  & 16 (13) & 34 (18) & 14 (11) & 36 (20) & 20 (11) & 30 (20) & 1854.12 (611.52) & 1805.5 (555.9) & 1931.72 (763.52) \\
        325 & 0.3685  & 13 (9) & 37 (11) & 10 (8) & 40 (12) & 3 (1) & 47 (19) & 18753.9 (4945) & 18788.72 (5244.9) & 18704.34 (5212.85) \\
        450 & 0.37  & 15 (14) & 35 (16) & 14 (14) & 36 (16) & 8 (7) & 42 (23) & 59911.7 (18365.47) & 61718.2 (21992) & 61375.54 (21728.17) \\
        600 & 0.3669  & 37 (9) & 13 (9) & 33 (8) & 17 (10) & 1 (1) & 49 (17) & 316711.3 (109317.78) & 316688.28 (104285.89) & 315321.52 (103800.06) \\
        800 & 0.366  & 5 (4) & 45 (4) & 3 (3) & 47 (5) & 0 (0) & 50 (8) & 1497034.74 (305983) & 1520004.3 (477415.5) & 1485407.12 (326572.38) \\
        \bottomrule
    \end{tabular}}
\end{table}

\section{Raw Experimental Results for Finding Minimum Dominating Cliques}

We train three GNN models for finding the minimum dominating clique. The loss function of the first GNN model is Equation (\ref{loss_for_dcs}). For the other two GNN models, both of their loss functions are Equation (\ref{loss_for_min_dcs}). The difference is that one calculates $\mathbb{E}(|S|)$ as $\sum_{i=1}^{n} p_i$ and another one calculates $\mathbb{E}(|S|)$ as Equation (\ref{subspace_expected}). With the fast and accurate versions of the joint information entropy, we have six GNN-enhanced solvers. We run these six solvers and the original solver with the new features on the training, validation and test data from the instances generated from $G(n,p)$ where $p \in (0.4, 0.41)$. The results are shown in Tables~\ref{tab:cmp_training_dataset_for_min}, \ref{tab:cmp_validation_dataset_for_min}, and \ref{tab:cmp_test_dataset_for_min}. The O in the tables represents the original solver with the new features. The (\ref{loss_for_dcs}) in the tables represents the solver using the probability distributions from the GNN model with Equation (\ref{loss_for_dcs}) as the loss function. The (\ref{loss_for_min_dcs}) in the tables represents the solver using the probability distributions from the GNN model with Equation (\ref{loss_for_min_dcs}) where $\mathbb{E}(|S|) =\sum_{i=1}^{n} p_i$ as the loss function, and the (\ref{subspace_expected}) represents the solver using the probability distributions from the GNN model whose loss function is Equation (\ref{loss_for_min_dcs}) where $\mathbb{E}(|S|)$ is calculated as Equation (\ref{subspace_expected}). For each pair of $n$ and $p$, there are two rows. The first row records the results of the solver using the fast version of the joint information entropy, and the second row records the results of the solver using the accurate version. 

\begin{table}
    \scriptsize
    \center
    \caption{Results on Training Data (minimum DC): 50 instances for each $n$.}
    \label{tab:cmp_training_dataset_for_min}
    \resizebox{1\textwidth}{!}{
    \begin{tabular}{ccccccccccccccllll}
        \toprule
        \multicolumn{2}{c}{$G(n,p)$} & \multicolumn{2}{c}{O vs (3)} & \multicolumn{2}{c}{O vs (4)} & \multicolumn{2}{c}{O vs (5)} & \multicolumn{2}{c}{(3) vs (4)} & \multicolumn{2}{c}{(3) vs (5)} & \multicolumn{2}{c}{(4) vs (5)} & \multicolumn{4}{c}{Avg} \\
        \cmidrule(lr){1-2} \cmidrule(lr){3-4} \cmidrule(lr){5-6} \cmidrule(lr){7-8} \cmidrule(lr){9-10} \cmidrule(lr){11-12} \cmidrule(lr){13-14} \cmidrule(lr){15-18}
        \(\text{$n$}\) & \(\text{$p$}\) & \(\mathbf{O}\) & \(\mathbf{(3)}\) & \(\mathbf{O}\) & \(\mathbf{(4)}\) & \(\mathbf{O}\) & \(\mathbf{(5)}\) & \(\mathbf{(3)}\) & \(\mathbf{(4)}\) & \(\mathbf{(3)}\) & \(\mathbf{(5)}\) & \(\mathbf{(4)}\) & \(\mathbf{(5)}\) & \(\mathbf{O}\) & \(\mathbf{(3)}\) & \(\mathbf{(4)}\) & \(\mathbf{(5)}\) \\
        \midrule
        75 & 0.4045  & 22 & 28 & 24 & 26 & 24 & 26 & 21 & 29 & 22 & 28 & 23 & 27 & 308.68 & 306.7 & 306.24 & 307.2 \\
        & & 18 & 32 & 22 & 28 & 22 & 28 & 27 & 23 & 24 & 26 & 21 & 29 &  & 303.8 & 306.06 & 305.4 \\

        150 & 0.4047  & 23 & 27 & 20 & 30 & 19 & 31 & 18 & 32 & 16 & 34 & 16 & 34 & 3637.72 & 3631.3 & 3618.64 & 3602.6 \\
        & & 14 & 36 & 16 & 34 & 10 & 40 & 23 & 27 & 16 & 34 & 16 & 34 &  & 3604.88 & 3599.78 & 3569.96 \\

        225 & 0.4095  & 18 & 32 & 15 & 35 & 14 & 36 & 21 & 29 & 16 & 34 & 24 & 26 & 22602.64 & 22601.52 & 22491.72 & 22479.42 \\
        & & 10 & 40 & 8 & 42 & 5 & 45 & 24 & 26 & 14 & 36 & 19 & 31 &  & 22509.86 & 22373.34 & 22319.28 \\

        250 & 0.405  & 20 & 30 & 13 & 37 & 13 & 37 & 16 & 34 & 15 & 35 & 21 & 29 & 31677.9 & 31708.42 & 31563.08 & 31537.88 \\
        & & 6 & 44 & 6 & 44 & 2 & 48 & 21 & 29 & 19 & 31 & 19 & 31 &  & 31468.26 & 31401.42 & 31372.74 \\

        300 & 0.4033  & 14 & 36 & 8 & 42 & 7 & 43 & 19 & 31 & 14 & 36 & 14 & 36 & 67677.04 & 67415.94 & 67256.94 & 67129.94 \\
        & & 3 & 47 & 2 & 48 & 2 & 48 & 23 & 27 & 14 & 36 & 17 & 33 &  & 66917.22 & 66842.58 & 66807.16 \\

        350 & 0.4016  & 5 & 45 & 4 & 46 & 2 & 48 & 27 & 23 & 15 & 35 & 15 & 35 & 131360.72 & 130768.52 & 130718.28 & 130617.76 \\
        & & 0 & 50 & 1 & 49 & 3 & 47 & 27 & 23 & 23 & 27 & 23 & 27 &  & 129634.98 & 129750.82 & 129822.26 \\

        375 & 0.4095 & 7 & 43 & 6 & 44 & 5 & 45 & 16 & 34 & 8 & 42 & 16 & 34 & 245800.8 & 244668.44 & 244319.66 & 244151.9 \\
        & & 3 & 47 & 3 & 47 & 3 & 47 & 22 & 28 & 14 & 36 & 17 & 33 &  & 242765.78 & 242628.48 & 242606.26 \\

        425 & 0.4085 & 9 & 41 & 3 & 47 & 2 & 48 & 22 & 28 & 9 & 41 & 10 & 40 & 435390.16 & 433530.2 & 433293.12 & 432740.06 \\
        & & 0 & 50 & 0 & 50 & 0 & 50 & 28 & 22 & 19 & 31 & 15 & 35 &  & 429838.34 & 429872.68 & 429567.34 \\

        475 & 0.4085 & 12 & 38 & 7 & 43 & 3 & 47 & 14 & 36 & 3 & 47 & 15 & 35 & 769454.38 & 767403.16 & 766112.54 & 764304.22 \\
        & & 1 & 49 & 1 & 49 & 1 & 49 & 18 & 32 & 8 & 42 & 23 & 27 &  & 760701.14 & 760309.4 & 759021 \\

        525 & 0.4031 & 10 & 40 & 2 & 48 & 1 & 49 & 11 & 39 & 8 & 42 & 19 & 31 & 1008488.18 & 1005927.8 & 1004287.22 & 1003163.38 \\
        & & 1 & 49 & 2 & 48 & 2 & 48 & 13 & 37 & 18 & 32 & 30 & 20 &  & 997220.98 & 996463.64 & 996105.24 \\

        575 & 0.4041 & 3 & 47 & 1 & 49 & 1 & 49 & 1 & 49 & 4 & 46 & 30 & 20 & 1697368.32 & 1693151.32 & 1688889.7 & 1689307.82 \\
        & & 0 & 50 & 0 & 50 & 0 & 50 & 4 & 46 & 20 & 30 & 39 & 11 &  & 1677423.02 & 1675428.62 & 1676514.82 \\

        625 & 0.4044 & 3 & 47 & 3 & 47 & 5 & 45 & 13 & 37 & 7 & 43 & 7 & 43 & 2676091.54 & 2670711.66 & 2668825.26 & 2664485.52 \\
        & & 0 & 50 & 0 & 50 & 1 & 49 & 18 & 32 & 9 & 41 & 11 & 39 &  & 2646282.86 & 2645545.64 & 2644130.92 \\

        650 & 0.407 & 5 & 45 & 1 & 49 & 4 & 46 & 8 & 42 & 9 & 41 & 10 & 40 & 3719504.28 & 3710680.98 & 3707591.22 & 3702165.74 \\
        & & 0 & 50 & 0 & 50 & 0 & 50 & 19 & 31 & 13 & 37 & 18 & 32 &  & 3675161.72 & 3673243.42 & 3672913.5 \\

        675 & 0.4057 & 7 & 43 & 2 & 48 & 7 & 43 & 5 & 45 & 8 & 42 & 35 & 15 & 4288894.48 & 4283315.68 & 4274585.02 & 4283440.86 \\
        & & 0 & 50 & 1 & 49 & 2 & 48 & 15 & 35 & 20 & 30 & 31 & 19 &  & 4242175.36 & 4237965.16 & 4245826.02 \\

        700 & 0.4012 & 8 & 42 & 6 & 44 & 6 & 44 & 14 & 36 & 18 & 32 & 24 & 26 & 4246975.12 & 4239157.58 & 4235418.36 & 4235944.28 \\
        & & 1 & 49 & 0 & 50 & 1 & 49 & 13 & 37 & 18 & 32 & 20 & 30 &  & 4202792.04 & 4198752.54 & 4198517.02 \\
        \bottomrule
    \end{tabular}}
\end{table}

\begin{table}[H]
    \scriptsize
    \center
    \caption{Results on Validation Data (minimum DC): 50 instances for each $n$.}
    \label{tab:cmp_validation_dataset_for_min}
    \resizebox{1\textwidth}{!}{
    \begin{tabular}{ccccccccccccccllll}
        \toprule
        \multicolumn{2}{c}{$G(n,p)$} & \multicolumn{2}{c}{O vs (3)} & \multicolumn{2}{c}{O vs (4)} & \multicolumn{2}{c}{O vs (5)} & \multicolumn{2}{c}{(3) vs (4)} & \multicolumn{2}{c}{(3) vs (5)} & \multicolumn{2}{c}{(4) vs (5)} & \multicolumn{4}{c}{Avg} \\
        \cmidrule(lr){1-2} \cmidrule(lr){3-4} \cmidrule(lr){5-6} \cmidrule(lr){7-8} \cmidrule(lr){9-10} \cmidrule(lr){11-12} \cmidrule(lr){13-14} \cmidrule(lr){15-18}
        \(\text{$n$}\) & \(\text{$p$}\) & \(\mathbf{O}\) & \(\mathbf{(3)}\) & \(\mathbf{O}\) & \(\mathbf{(4)}\) & \(\mathbf{O}\) & \(\mathbf{(5)}\) & \(\mathbf{(3)}\) & \(\mathbf{(4)}\) & \(\mathbf{(3)}\) & \(\mathbf{(5)}\) & \(\mathbf{(4)}\) & \(\mathbf{(5)}\) & \(\mathbf{O}\) & \(\mathbf{(3)}\) & \(\mathbf{(4)}\) & \(\mathbf{(5)}\) \\
        \midrule
        100 & 0.4006 & 24 & 26 & 24 & 26 & 20 & 30 & 27 & 23 & 22 & 28 & 21 & 29 & 745.38 & 746.28 & 744.36 & 746.14 \\
        & & 17 & 33 & 21 & 29 & 15 & 35 & 24 & 26 & 17 & 33 & 17 & 33 &  & 742.5 & 741.6 & 741.14 \\

        275 & 0.4023 & 14 & 36 & 14 & 36 & 7 & 43 & 23 & 27 & 13 & 37 & 13 & 37 & 44691.44 & 44530.68 & 44548.94 & 44327.94 \\
        & & 1 & 49 & 4 & 46 & 4 & 46 & 21 & 29 & 17 & 33 & 16 & 34 &  & 44219.1 & 44261.68 & 44110.74 \\

        400 & 0.4018 & 9 & 41 & 7 & 43 & 2 & 48 & 21 & 29 & 8 & 42 & 11 & 39 & 245225.78 & 244438.76 & 244364.86 & 243192.42 \\
        & & 0 & 50 & 0 & 50 & 1 & 49 & 21 & 29 & 17 & 33 & 17 & 33 &  & 242227.96 & 242320.04 & 241723.42 \\

        550 & 0.407 & 3 & 47 & 2 & 48 & 1 & 49 & 4 & 46 & 1 & 49 & 17 & 33 & 1533702.56 & 1530039.08 & 1527699.04 & 1526218.12 \\
        & & 0 & 50 & 1 & 49 & 1 & 49 & 16 & 34 & 17 & 33 & 24 & 26 &  & 1515756.72 & 1515495.92 & 1514851 \\

        725 & 0.4045 & 7 & 43 & 9 & 41 & 0 & 50 & 37 & 13 & 4 & 46 & 0 & 50 & 5944743 & 5933028.72 & 5942993.54 & 5905492.66 \\
        & & 0 & 50 & 2 & 48 & 0 & 50 & 35 & 15 & 5 & 45 & 0 & 50 &  & 5879623.24 & 5890517.12 & 5861546.88 \\

        \bottomrule
    \end{tabular}}
\end{table}

\begin{table}[H]
    \scriptsize
    \center
    \caption{Results on Test Data (minimum DC): 50 instances for each $n$.}
    \label{tab:cmp_test_dataset_for_min}
    \resizebox{1\textwidth}{!}{
    \begin{tabular}{ccccccccccccccllll}
        \toprule
        \multicolumn{2}{c}{$G(n,p)$} & \multicolumn{2}{c}{O vs (3)} & \multicolumn{2}{c}{O vs (4)} & \multicolumn{2}{c}{O vs (5)} & \multicolumn{2}{c}{(3) vs (4)} & \multicolumn{2}{c}{(3) vs (5)} & \multicolumn{2}{c}{(4) vs (5)} & \multicolumn{4}{c}{Avg} \\
        \cmidrule(lr){1-2} \cmidrule(lr){3-4} \cmidrule(lr){5-6} \cmidrule(lr){7-8} \cmidrule(lr){9-10} \cmidrule(lr){11-12} \cmidrule(lr){13-14} \cmidrule(lr){15-18}
        \(\text{$n$}\) & \(\text{$p$}\) & \(\mathbf{O}\) & \(\mathbf{(3)}\) & \(\mathbf{O}\) & \(\mathbf{(4)}\) & \(\mathbf{O}\) & \(\mathbf{(5)}\) & \(\mathbf{(3)}\) & \(\mathbf{(4)}\) & \(\mathbf{(3)}\) & \(\mathbf{(5)}\) & \(\mathbf{(4)}\) & \(\mathbf{(5)}\) & \(\mathbf{O}\) & \(\mathbf{(3)}\) & \(\mathbf{(4)}\) & \(\mathbf{(5)}\) \\
        \midrule
        200 & 0.406 & 19 & 31 & 16 & 34 & 11 & 39 & 20 & 30 & 13 & 37 & 20 & 30 & 12762.36 & 12745.5 & 12710.6 & 12668.88 \\
        & & 7 & 43 & 8 & 42 & 3 & 47 & 16 & 34 & 13 & 37 & 20 & 30 &  & 12664.72 & 12603.94 & 12610.1 \\

        325 & 0.4087 & 7 & 43 & 10 & 40 & 6 & 44 & 26 & 24 & 22 & 28 & 18 & 32 & 117555.52 & 116940.76 & 116912.58 & 116758.82 \\
        & & 3 & 47 & 3 & 47 & 3 & 47 & 30 & 20 & 24 & 26 & 20 & 30 &  & 116137.26 & 116220.76 & 116027.64 \\

        450 & 0.4075 & 5 & 45 & 8 & 42 & 4 & 46 & 11 & 39 & 4 & 46 & 10 & 40 & 561310.58 & 559585.54 & 558974.04 & 556734.04 \\
        & & 2 & 48 & 1 & 49 & 2 & 48 & 22 & 28 & 16 & 34 & 16 & 34 &  & 554707.44 & 554511.8 & 553003.66 \\

        600 & 0.4017 & 5 & 45 & 0 & 50 & 1 & 49 & 6 & 44 & 4 & 46 & 29 & 21 & 1921609.08 & 1917237.3 & 1912831.24 & 1912535.74 \\
        & & 0 & 50 & 0 & 50 & 0 & 50 & 6 & 44 & 20 & 30 & 39 & 11 &  & 1898861.26 & 1896397.86 & 1898055.04 \\

        800 & 0.4062 & 24 & 26 & 12 & 38 & 4 & 46 & 15 & 35 & 4 & 46 & 6 & 44 & 11202974.98 & 11206548.2 & 11195113.58 & 11161754.14 \\
        & & 2 & 48 & 0 & 50 & 3 & 47 & 33 & 17 & 4 & 46 & 5 & 45 &  & 11105365.36 & 11106311.6 & 11071003.9 \\
        \bottomrule
    \end{tabular}}
\end{table}

\end{document}